\documentclass{article}

\usepackage[backend=bibtex, sorting=none]{biblatex}
\usepackage[affil-it]{authblk}
\usepackage[pdftex]{graphicx}
\usepackage{amssymb}
\usepackage{amsmath}
\usepackage{amsthm}
\usepackage{amsfonts}
\usepackage{bm}
\usepackage{xcolor}

\DeclareMathOperator*{\argmin}{arg\,min}  

\title{Deep reinforcement learning uncovers processes for separating azeotropic mixtures without prior knowledge}

\author[1, $\dagger$]{Quirin Göttl}

\author[2, 3, $\dagger$]{Jonathan Pirnay}
\author[1, $\ddagger$, $\ast$]{Jakob Burger}
\author[2, 3, 4, $\ddagger$, $\ast$]{Dominik G. Grimm}
\affil[1]{Technical University of Munich, TUM Campus Straubing for Biotechnology and Sustainability, Laboratory of Chemical Process Engineering, Schulgasse 16, 94315 Straubing, Germany.}
\affil[2]{Technical University of Munich, TUM Campus Straubing for Biotechnology and Sustainability, Bioinformatics, Petersgasse 18, 94315 Straubing, Germany.}
\affil[3]{Weihenstephan-Triesdorf University of Applied Sciences, Petersgasse 18, 94315 Straubing, Germany.}
\affil[4]{Technical University of Munich, TUM School of Computation, Information and Technology (CIT), Boltzmannstraße 3, 85748 Garching, Germany.}
\affil[$\dagger$]{These authors contributed equally as joint first authors.}
\affil[$\ddagger$]{These authors contributed equally as joint last authors.}
\affil[$\ast$]{Corresponding authors: burger@tum.de, dominik.grimm@tum.de}

\date{\vspace{-5ex}}

\usepackage{hyperref}

\begin{document}
\maketitle

\begin{abstract}

Process synthesis in chemical engineering is a complex planning problem due to vast search spaces, continuous parameters and the need for generalization. Deep reinforcement learning agents, trained without prior knowledge, have shown to outperform humans in various complex planning problems in recent years. Existing work on reinforcement learning for flowsheet synthesis shows promising concepts, but focuses on narrow problems in a single chemical system, limiting its practicality. We present a general deep reinforcement learning approach for flowsheet synthesis. We demonstrate the adaptability of a single agent to the general task of separating binary azeotropic mixtures. Without prior knowledge, it learns to craft near-optimal flowsheets for multiple chemical systems, considering different feed compositions and conceptual approaches. On average, the agent can separate more than 99\% of the involved materials into pure components, while autonomously learning fundamental process engineering paradigms. This highlights the agent’s planning flexibility, an encouraging step toward true generality.

\end{abstract}

\section{Introduction}

Process systems engineering (PSE) in (bio-)chemical engineering is the development of systematic techniques for process modeling, design, and control \cite{sargent1983}. Current advances in data-driven methods have a significant impact on PSE \cite{venkatasubramanian2018, lee2018, dobbelaere2021, schweidtmann2021, pistikopoulos2021}, e.g., in the fields of computer-aided molecular design \cite{yang2019, coley2019, ross2022}, reaction engineering \cite{schwaller2021, zhou2017}, and process (surrogate) modeling \cite{mcbride2019}.

In advancing to a sustainable chemical industry, the contribution of machine learning (ML) is pivotal in minimizing resource consumption and waste generation. A critical application in this direction is conceptual process synthesis: given desired products and a set of available raw material streams, the aim is to find the process flowsheet that maximizes a predefined objective such as monetary profit. Traditionally, this task has relied on heuristic knowledge and manual design in combination with process simulation \cite{siirola1996, westerberg2004, montastruc2019}. First steps toward automated flowsheet synthesis (AFS) were marked by process simulation for evaluation of process alternatives and algorithmic expert systems \cite{kirkwood1988, gani1989}, later by optimization schemes that reduce superstructures to viable flowsheet options \cite{mencarelli2020}. Those approaches rely to a certain degree on guidance by humans, e.g., by providing heuristics or an algorithmic rule-set, as the space of possible flowsheets is enormous.

ML, particularly deep reinforcement learning (RL), enables a paradigm change. It has been shown that RL agents learning without prior expert knowledge can outperform humans in domains with combinatorial search spaces \cite{silver2017, silver2018, fawzi2022discovering, mankowitz2023faster}. Applied to AFS, an RL agent observes an initial situation (usually a set of raw material feed streams) and designs a flowsheet by sequentially placing units and connecting them with present ones. After each placement, a process simulator returns the current state of the flowsheet to the agent. Once the agent finishes the synthesis, the objective function is evaluated on the constructed flowsheet as a scalar reward. The agent's task is to maximize this reward while interacting with the process simulator only. Formulated in this generality, the complexity of the problem is enormous: not only is the agent's search space exponentially increased with each possible type of unit to choose from, but also the continuous specifications of the placed units transform the agent's action space from discrete into discrete-continuous hybrid. Furthermore, recycle loops within the process can drastically change the compositions and flowrates of earlier placed streams, so the agent needs to plan its actions for the long term. 
Complicating the challenge to the extreme, {\em generality} is the holy grail of all applied fields of artificial intelligence: a genuinely general agent should be capable of designing flowsheets for many tasks and generalizing beyond known situations and initial feed compositions.

Current work on reinforcement learning for AFS shows proof of concept with a narrow scope where the agent is trained on a single initial situation (i.e., a single feed stream composition) or within a simplified discrete action space. Midgley \cite{midgley2020} lets an agent set up sequences of (discrete) distillation columns to optimize a single problem instance. Khan et al. \cite{khan2020} train an agent to search for optimal process routes for hydrogen production with a simplified hybrid action space, yet also for a singular instance of the problem. G\"ottl et al. \cite{goettl2022a} pose AFS as a competitive two-player game to enforce exploration and evade reward shaping. They train the agent with Monte Carlo tree search (MCTS)-based self-play similar to AlphaZero \cite{silver2017,silver2018}. Although the agent's action space is simplified to be purely discrete, they train and evaluate the agent on random feed compositions of a single quaternary reaction-distillation problem. Their gamified approach is further improved by hierarchically structuring the agent's discrete action space \cite{goettl2021a, goettl2021b, goettl2022b}. Khan et al. \cite{khan2022} also employ a hierarchical action space for an agent learning to construct a process for ethylene oxide production. Seidenberg et al. \cite{seidenberg2023} continue this work where an ontological framework supports the agent's decisions. Stops et al. \cite{stops2023} construct a process for producing methyl acetate, where a graph neural network (GNN) for the agent's policy naturally represents a flowsheet's structure. They train and evaluate the policy network with the PPO algorithm \cite{schulman2017} on a fixed feed composition. That work is continued in \cite{gao2023} using a form of transfer learning, meaning that the agent was pre-trained on a simulation employing short-cut models and fine-tuned with a rigorous process simulation. The mentioned approaches consider single or randomly generated problem instances (feed compositions) during training, but all constrain the agent to a single problem class, i.e., one chemical system. 

Yet the first step to generalization is flexibility. Hence, in this work, we take a step towards generalizing AFS through deep RL by showing for the first time that a single agent can learn to synthesize near-optimal flowsheets for {\em multiple} chemical systems, each with varying feed compositions, each requiring a substantially different conceptual approach. As an example task, due to its complex nature, we take the separation of azeotropic mixtures and train an agent from zero knowledge to design flowsheets for multiple chemical systems. The agent learns by only interacting with a process simulator, using an algorithmic further development of AlphaZero \cite{silver2018,danihelka2022}. We introduce a tree-pruning technique that enables the agent to learn when the process simulator diverges, and the flowsheet fails – a known challenge in deep RL for AFS that has not been addressed satisfactorily so far. We propose to view flowsheet encoding as a sequence-to-sequence problem and use the multilayer perceptron (MLP)-based MLP-Mixer \cite{tolstikhin2021mlp} – diverted from computer vision – as the base architecture for the policy and value neural network. It can distill the dynamics of the environment such that the policy network alone can produce excellent flowsheets.

Furthermore, to efficiently navigate the search tree, we structure the hybrid action space hierarchically into multiple levels, ranging from selecting an open stream and a unit to continuous specifications of the placed unit. By factorizing a discretization of these continuous parameters and including them in the hierarchy, the agent quickly learns to set meaningful unit specifications. Additionally, we show the capability of the network architecture to generate flowsheets by unrolling the policy alone with beam search, a breadth-first search of limited width, to obtain action sequences with high total probability.

The agent is trained simultaneously on several feed compositions from four chemical systems: acetone-chloroform, ethanol-water, butanol-water, and pyridine-water. It learns to add solvents on demand, combine them with distillation columns, decanters, and mixers, and place crucial recycle loops. Without prior knowledge, it discovers azeotropic and entrainer distillation, two classical chemical engineering schemes. In 78.5\% of all tested instances, the trained agent constructs a flowsheet separating the feed into pure streams and recovering used solvents entirely. Additionally, our analysis shows that the proposed flowsheets separate, on average, 99\% of the feed and added solvent into pure streams. Therefore, by showing in this work that a single agent can flexibly solve a multitude of problems, we mark a significant step towards true generality, i.e., an agent that can transfer its learnings from the training process to conceptual design problems it has yet to encounter.

\section{Results}

\subsection{AlphaZero for general flowsheet synthesis}

We pose AFS as a single-agent RL problem \cite{sutton2018reinforcement} and represent each state $s_t$ at a timestep $t$ in the environment primarily by a matrix $\bm M_{s_t}$ in which every nonzero line describes a stream and its connections to other streams of the currently designed flowsheet. At timestep $t$, the agent receives the current flowsheet matrix $\bm M_{s_t}$ and fills the remaining lines sequentially by placing units on the flowsheet. It executes hierarchically structured actions $a_t$, which consist of choosing an open stream, picking a unit from a predefined pool, and further specifying the unit – if required – by setting a continuous parameter or stream destination (when mixing or recycling streams). In this work, the agent has to set the following continuous specifications: the amount of solvent to add to a stream and the ratio of distillate to feed flowrates inside a distillation column. The synthesis ends when the agent executes a special termination action, or when the flowsheet matrix is full. Hence, by limiting the number of matrix lines, the agent is naturally forced to look for shorter solutions. Once the synthesis terminates, a reward is obtained that depends on a predefined objective function, and the agent's goal is to maximize this reward.

Predominantly, we let the agent learn with the algorithmic Gumbel extension \cite{danihelka2022} of AlphaZero \cite{silver2018}, where at each timestep, the agent performs an MCTS to choose its subsequent hierarchical action. A neural network guides the MCTS, which takes as input the flowsheet matrix $\bm M_{s_t}$ and information about already decided specifications for the next unit to place. The network outputs a scalar value function estimation and a policy prediction. The value estimates the expected final reward the agent obtains when it continues its design from $s_t$. The policy prediction is a probability distribution over the actions at the current state. The flowsheets designed by the agent and the search tree statistics are used to train the neural network in a supervised manner. 

\subsection{Neural network architecture}

\begin{figure}
	\centering
	\includegraphics[width=\linewidth]{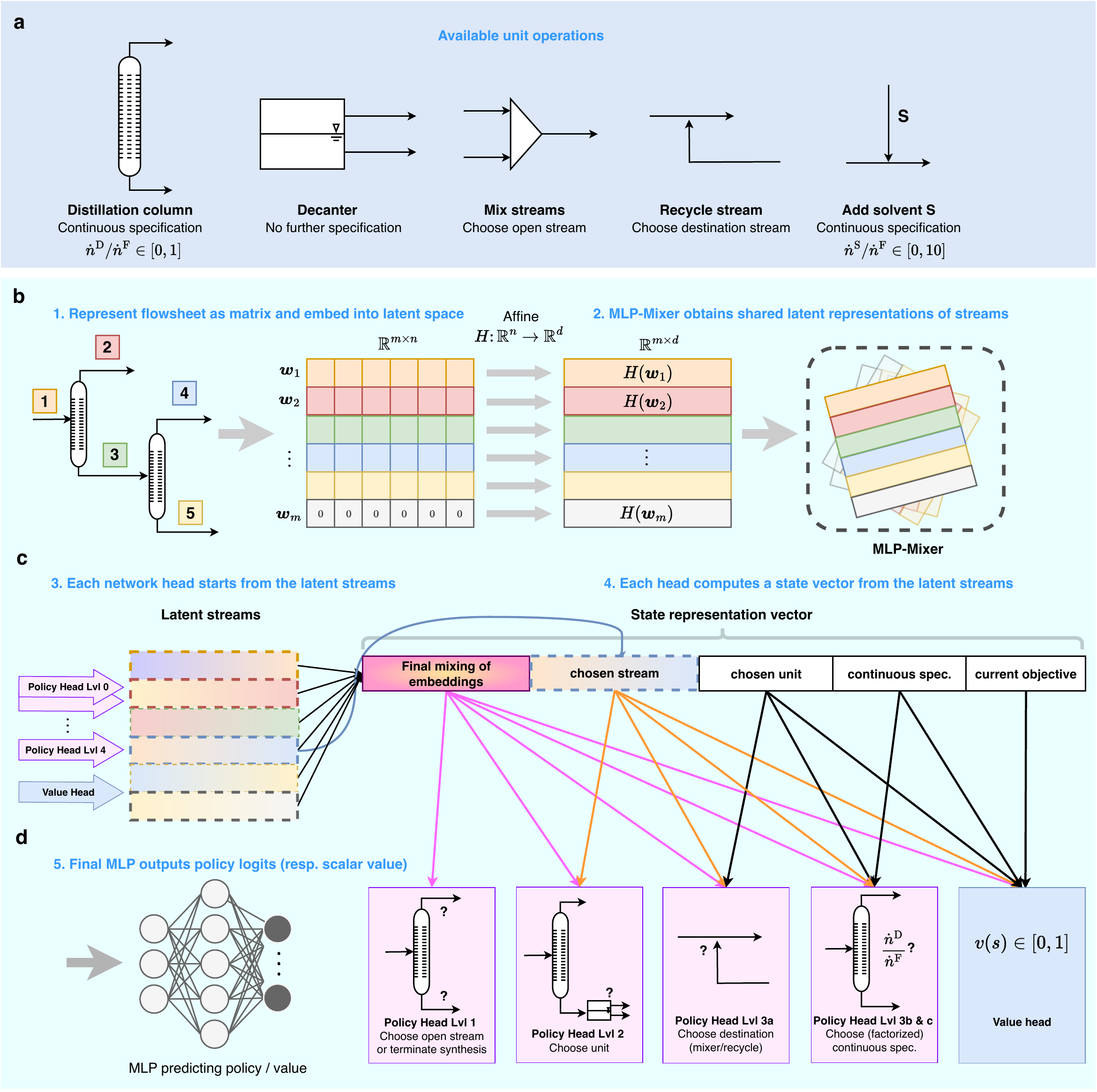}
	\caption{{\bf Method overview. }{\bf a,} A design decision consists of placing a parameterized unit on an open stream. The agent must set continuous parameters for the distillation column (ratio of distillate flowrate to feed flowrate) and when adding a solvent (ratio of solvent flowrate to the flowrate of chosen stream). The agent must pick a second stream from the flowsheet when mixing or recycling streams. {\bf b,} We compute a latent representation of the flowsheet streams by first encoding the current flowsheet as a sequence of vectors $\bm w_1, \dots, \bm w_m$. The affinely embedded sequence is passed to an MLP-Mixer, transforming the sequence into latent streams. {\bf c,} The latent streams are shared with separate network heads for each hierarchy level of the policy, as well as a separate head for the value prediction. Each head combines the latent streams to a final vector and enriches it with information from previous hierarchy levels. {\bf d,} The state vector is passed to a final MLP predicting the probability distribution over the hierarchical level (in the case of a policy head) or outputting a scalar value estimation. We summarize the network heads on the right-hand side, with arrows from the state vector indicating which information is passed to which head.}
	\label{figure_method}
\end{figure}

AlphaZero-type algorithms stand and fall with their underlying neural network. In our case, it must provide the agent with latent embeddings of the flowsheet streams that capture the essence of the current state $s_t$ and allow the agent to derive accurate policy and value predictions. Hence, the network always needs to maintain a global view of all streams and their connections to each other, as choosing to recycle a stream can alter all present streams simultaneously. Convolutional networks and GNNs generally struggle to capture long-range dependencies. In particular, we pose the task of obtaining stream representations as a sequence-to-sequence problem: Let  $\bm M_s \in \mathbb R^{m \times n}$ be the flowsheet matrix at a state $s$ with lines $\bm w_1, \dots, \bm w_m$ and affine embeddings $H(\bm w_i) \in \mathbb R^d$ into some latent space of dimension $d$. We propose to use a neural network based on the MLP-Mixer \cite{tolstikhin2021mlp} to transform the sequence of matrix lines $\bm w_1, \dots, \bm w_m$ into an expressive sequence of latent stream embeddings $\text{MLPMixer}(H(\bm w_1), \dots, H(\bm w_m)) \in \mathbb R^{m \times d}$. Although the MLP-Mixer might seem an unorthodox choice as its primary use is classically in computer vision, it has a global receptive field (as in self-attention based transformers \cite{vaswani2017}), but with only linear complexity in the number of matrix lines (as opposed to the quadratic complexity of transformers). Furthermore, it reflects the sequential procedure of placing unit after unit on the flowsheet. We provide an overview of our method in Figure~\ref{figure_method}.

\subsection{Problem setup}

The agent's task is to separate a binary feed stream into its pure components, where we denote by $x_1 \in (0,1)$ the molar fraction of the first component and by $x_2 = 1-x_1$ the molar fraction of the second component. We consider four chemical example systems from the literature \cite{wang2018, kunnakorn2013, luyben2008, chen2015}, as listed in Table \ref{table_example_systems}. The available unit operations (displayed in Figure \ref{figure_method}) are simulated using short-cut models \cite{ryll2012a, ryll2012b, goettl2023}, which rely on material balances and thermodynamic limits, assuming the best possible performance. Details are provided in Section \ref{method_section_environment}. 

\begin{table}
	\caption{Considered chemical example systems together with the available solvents. We use the following abbreviations for the components: acetone (Ac), benzene (Be), butanol (Bu), chloroform (Ch), ethanol (Et), pyridine (Py), tetrahydrofuran (Te), toluene (To), water (Wa).}
	\label{table_example_systems}
	\begin{center}
		\begin{tabular}{|p{4cm}||c|c|c|c|}
			\hline
			& System 1 & System 2 & System 3 & System 4 \\
			\hline
			Feed stream components & Ac, Ch & Et, Wa & Bu, Wa & Py, Wa \\
			\hline
			Available solvents & Be, To & Be, To, Te & Ac, Be, To & To \\
			\hline
			Temperature for decanter & 323.15 K & 338.15 K & 338.15 K & 338.15 K \\
			\hline
			Pressure for distillation, decanter & 1 bar & 1 bar & 1 bar & 1 bar \\
			\hline
		\end{tabular}
	\end{center}
\end{table}

\begin{figure}[t!]
	\includegraphics[width=\linewidth]{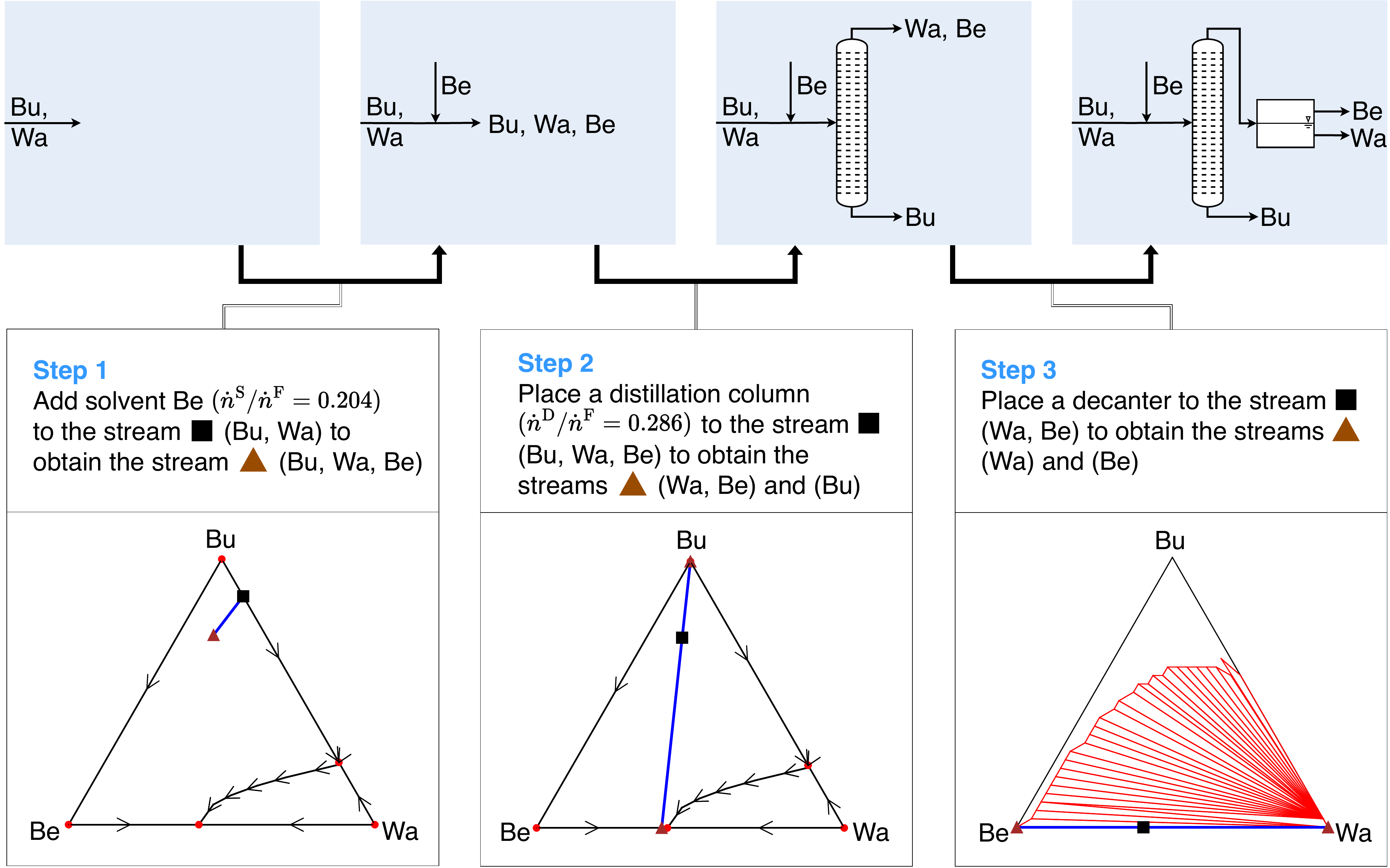}
	\caption{Stepwise construction process of a flowsheet for system 3 (feed composition: $x_{\mathrm{Bu}}=0.86, x_{\mathrm{Wa}}=0.14$) proposed by the trained agent. The lower panels visualize the ternary molar composition space together with distillation boundaries (Step 1 and Step 2) and liquid-liquid equilibria (Step 3) at the pressure and temperature conditions listed in Table \ref{table_example_systems}. In every step, the feed stream of the respective unit is indicated by a $\blacksquare$, and a brown $\blacktriangle$ marks the resulting output stream(s). In Step 1, the agent adds the solvent benzene to the feed, and as can be seen in the ternary diagram, this shifts the binary feed along the blue line into the ternary space. In Step 2,  a distillation column separates the ternary feed into a water -- benzene mixture and pure butanol (the blue line in the ternary diagram shows the split). Step 3 uses a decanter to split the water -- benzene mixture into highly pure components, which is possible as this system displays a liquid phase split visualized by the red triangles in the ternary diagram. After Step 3, the agent chooses to terminate the synthesis as all streams that leave the process are pure, and therefore, the separation task is completed. We explain ternary diagrams, phase equilibria, and distillation boundaries within the Supplementary Material \ref{supp_section_flowsheet_simulation}.}
	\label{figure_schrittweiser_aufbau}
\end{figure}

For a better understanding, Figure \ref{figure_schrittweiser_aufbau} shows the stepwise construction process of a flowsheet for system 3 as proposed by the trained agent.

From a chemical engineering viewpoint, solving the separation tasks requires conceptually quite different approaches. For example, mixtures in system 1 consist of acetone and chloroform and can be treated by entrainer distillation \cite{wang2018}. In particular, solvents benzene or toluene can be added as an entrainer to employ the curvature of the resulting ternary distillation boundaries to separate the feed stream. On the contrary, mixtures in system 2 consist of ethanol and water and can be separated using (heterogeneous) azeotropic distillation \cite{kunnakorn2013}. Hence, adding a solvent such as benzene results in a ternary mixture with liquid phase splits, which can be used to overcome the azeotrope. With minor modifications, these techniques can also be used for system 3 (as shown in Figure \ref{figure_schrittweiser_aufbau}) and system 4, but there are other options. For example, feed streams in system 3 can be separated without a solvent.

During training, the agent encounters 50k randomly sampled feed compositions (details on the sampling process are provided within the Supplementary Material \ref{supp_section_sampling}). We train the agent separately with two versions of the reward function, $\mathcal{C}_{\textnormal{\footnotesize literature}}$ and $\mathcal{C}_{\textnormal{\footnotesize generic}}$, respectively, to show the flexibility within the algorithmic framework. Reward $\mathcal{C}_{\textnormal{\footnotesize literature}}$ is based on literature about the economical analysis and cost of chemical processes. Reward $\mathcal{C}_{\textnormal{\footnotesize generic}}$ aims at providing a simple cost function for separation processes. It is handy for conceptual flowsheet synthesis, where the primary goal is not to obtain the most profitable solution but a feasible one. Certainly, modifying or replacing the reward function is possible. One could, for example, introduce additional objectives such as the reduction of emissions to reflect sustainability aims. As with actor-critic methods, the success of AlphaZero's MCTS strongly relies on the quality of the value estimations \cite{van2016learning,pohlen2018observe,pirnay2023}, so we normalize both $\mathcal{C}_{\textnormal{\footnotesize literature}}$ and $\mathcal{C}_{\textnormal{\footnotesize generic}}$ by the profit obtained in a perfect separation without any costs, and clip below zero, such that the observed reward lies in $[0, 1]$ (where $1$ is an unattainable upper bound). As the values of $\mathcal{C}_{\textnormal{\footnotesize literature}}$ and $\mathcal{C}_{\textnormal{\footnotesize generic}}$ may be hard to interpret, we define a performance ratio $\mathcal{R}\in[0, 1]$ as a metric to evaluate a flowsheet. The metric $\mathcal{R}$ describes how much of the input (feed and added solvent) the flowsheet separates into pure components. As the main goal is a perfect separation into pure components, a value of $\mathcal{R}$ close to $1$ is desirable. Details on $\mathcal{C}_{\textnormal{\footnotesize literature}}$, $\mathcal{C}_{\textnormal{\footnotesize generic}}$, and $\mathcal{R}$ are provided in Section \ref{method_section_environment}. Finally, we evaluate the agent on a test set containing 49 feed stream situations for every chemical system from Table \ref{table_example_systems}. Here, in the $i$-th situation of a binary system, the molar fraction of the first component is set to $x_1=0.02\cdot i$. This way, evaluation of the agent on the whole range of the binary composition space is ensured.

For each action to take, we grant the agent 200 simulations within the MCTS for both training and testing. We emphasize that the agent is generally allowed to choose any unit and specification without restriction from the provided pool of units. We only constrain the agent to add a solvent once per problem instance (in all considered examples, it is enough to add at most one solvent to separate the feed stream). 

\subsection{Overall performance}

Table \ref{table_performance_ratio} reports the agent's performance ratio $\mathcal R$ on the test set covering the full range of molar fractions. Row 1 and 2 show results for the agent using MCTS. On average, the agent achieves $\mathcal R > 0.95$ for all considered chemical systems. In over 60\% of all test instances, the agent proposes a flowsheet separating the feed and added solvent entirely into pure streams. While the agent performs almost perfectly in some cases, it becomes clear that its performance differs from system to system, e.g., when comparing system 3 and system 4. A reason for this is the varying difficulty of the separation task. Feeds from system 3 can often be separated with fewer units or without the usage of recycles, as shown in Figure \ref{figure_schrittweiser_aufbau}. 
On the contrary, the location of the binary azeotropes in the other systems often requires more sophisticated flowsheet designs. A reason for the difference in $\mathcal R$ w.r.t. the choice of the reward function is that the specification of a pure stream differs from $\mathcal{C}_{\textnormal{\footnotesize literature}}$ (mass fraction greater than 0.99) to $\mathcal{C}_{\textnormal{\footnotesize generic}}$ (molar fraction greater than 0.99). Still, the agent consistently proposes flowsheets that separate large parts of the feed and added solvent into pure streams with $\mathcal R > 0.9$.

We further assess whether the agent's performance relies strongly on the computationally expensive MCTS when designing a flowsheet or whether the neural network alone can independently capture most flowsheet dynamics. To achieve this, we discard the value network and let the agent use the policy network alone. If the network can truly grasp the underlying thermodynamics, then a flowsheet stemming from an action sequence with high total probability should yield a high outcome. We unroll the policy with {\em beam search}, the de-facto standard sequence decoding method in natural language processing to obtain a set of high-probability sequences from the model. Beam search is a pruned breadth-first search of limited width $k$, where at each timestep, we expand the (maximum of) $k$ actions that lead to sequences with the highest total probability. Row 3 and 4 in Table \ref{table_performance_ratio} show the agent's results using a moderate beam width of $k = 512$, yielding a simpler yet wider search than MCTS. As can be seen, the agent can now master all situations with an almost perfect performance ratio. Additionally, the number of cases with complete separation increases for all cases. 

These results show the suitability of the MLP-Mixer architecture for flowsheet representation. Furthermore, beam search allows generating high-quality candidate flowsheets fast (in contrast to slower MCTS), which has practical advantages, for example, when an agent proposes the conceptual design of a flowsheet that serves as an initialization for process optimization.

\begin{table}
	\caption{Performance of the agent on the test set for both reward functions $\mathcal{C}_{\textnormal{\footnotesize literature}}$ and $\mathcal{C}_{\textnormal{\footnotesize generic}}$. The ratio $\mathcal R$ indicates how much of the input (feed and added solvent) the agent's flowsheet separates into pure components. Additionally, we report how often the agent proposes a flowsheet that separates the feed and added solvent completely into pure streams ('Compl. sep.'). Row 1 and 2 show the results for the agent using MCTS. Row 3 and 4 show the results for unrolling the policy with beam search (BS) of width 512.}
	\label{table_performance_ratio}
	\begin{center}	
		\begin{tabular}{|p{3cm}|c||c|c|c|c|c|}
			\hline
			& & All sys. & Ac, Ch & Et, Wa & Bu, Wa & Py, Wa \\
			\hline
			$\mathcal{C}_{\textnormal{\footnotesize literature}}$ (MCTS) & $\mathcal R$ & 95.9\% & 97.5\% & 97.0\% & 97.5\% & 91.6\% \\
			& Compl. sep. & 60.5\% & 84.0\% & 50.0\% & 84.0\% & 24.0\% \\
			\hline
			\hline
			$\mathcal{C}_{\textnormal{\footnotesize generic}}$ (MCTS) & $\mathcal R$ & 97.4\% & 98.5\% & 94.5\% & 98.8\% & 97.8\% \\
			& Compl. sep. & 65.5\% & 70.0\% & 30.0\% & 98.0\% & 64.0\% \\
			\hline
			\hline
			$\mathcal{C}_{\textnormal{\footnotesize literature}}$ (BS) & $\mathcal R$ & 98.9\% & 99.5\% & 98.1\% & 99.6\% & 98.5\% \\
			& Compl. sep. & 77.0\% & 94.0\% & 68.0\% & 88.0\% & 58.0\% \\
			\hline
			\hline
			$\mathcal{C}_{\textnormal{\footnotesize generic}}$ (BS) & $\mathcal R$ & 99.0\% & 99.3\% & 97.4\% & 100.0\% & 99.4\% \\
			& Compl. sep. & 78.5\% & 86.0\% & 60\% & 98.0\% & 70.0\% \\
			\hline
		\end{tabular}
	\end{center}
\end{table}

\subsection{Uncovering prominent designs}

\begin{figure}[t!]
	\centering
	\includegraphics[width=\linewidth]{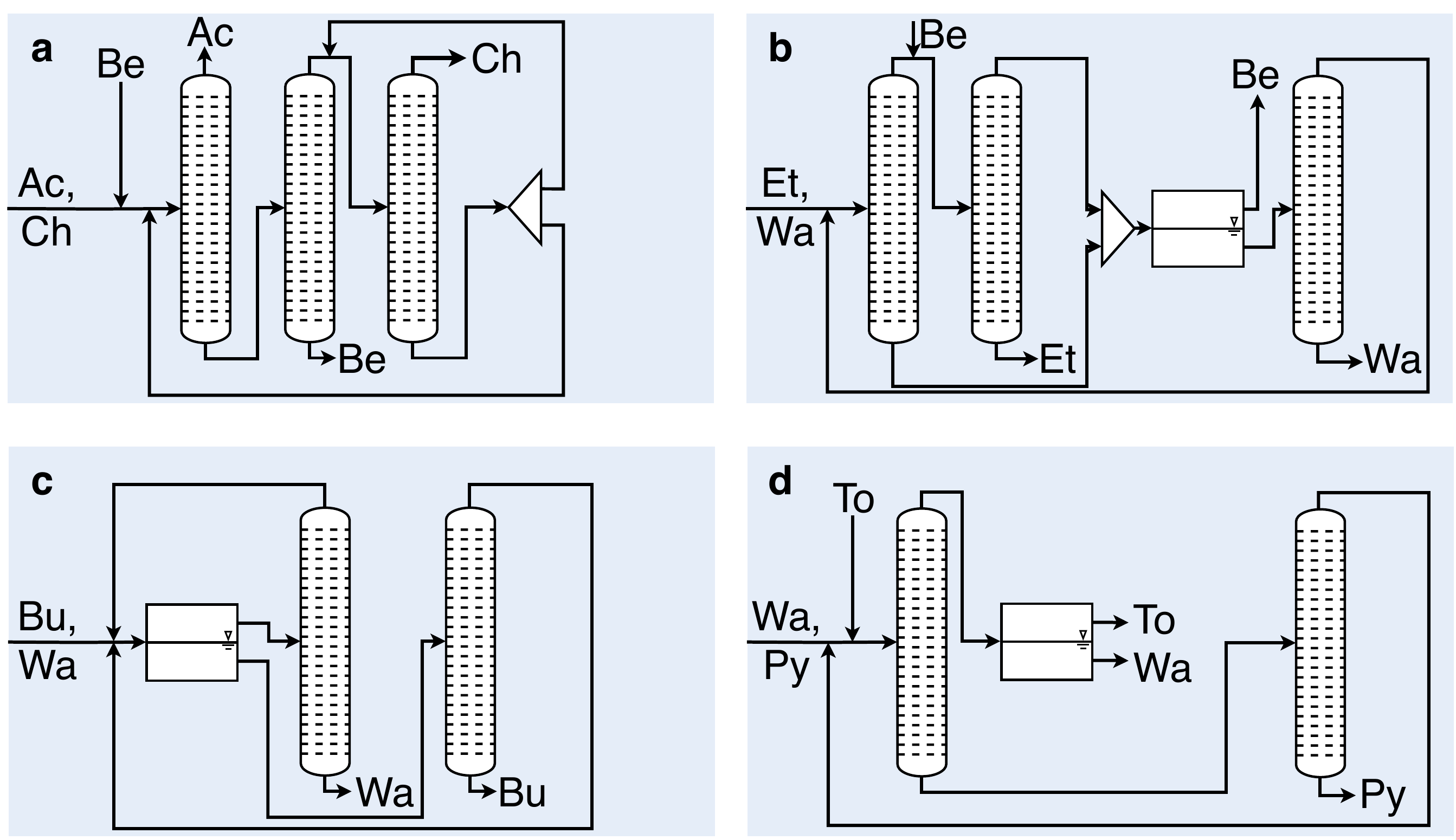}
	\caption{
	Flowsheets constructed by the trained agent for feed situations given in the literature \cite{wang2018, kunnakorn2013, luyben2008, chen2015} (training process was carried out using reward $\mathcal{C}_{\textnormal{\footnotesize literature}}$). {\bf a}, Separation of acetone and chloroform (feed composition: $x_{\mathrm{Ac}}=0.5, x_{\mathrm{Ch}}=0.5$) in an entrainer distillation using benzene as solvent. The agent adds benzene as an entrainer to the acetone -- chloroform feed to employ the resulting curvature of the ternary distillation boundaries to separate the streams. In this ternary system, there is no liquid phase split. Therefore, using a decanter does not help the separation (contrary to the other systems, e.g., as shown in Figure \ref{figure_schrittweiser_aufbau}). {\bf b}, Separation of ethanol and water (feed composition: $x_{\mathrm{Et}}=0.5, x_{\mathrm{Wa}}=0.5$) in an azeotropic distillation using solvent benzene. Both in {\bf b} and {\bf d}, a solvent is added that forms binary azeotropes with both feed components and is immiscible with water so that separation can be achieved by using a decanter in combination with a distillation sequence. {\bf c}, Azeotropic distillation of butanol and water (feed composition: $x_{\mathrm{Bu}}=0.4, x_{\mathrm{Wa}}=0.6$) without using a solvent. The reason for this is that the binary system butanol -- water already displays a liquid phase split, which allows the immediate usage of a decanter. {\bf d}, Azeotropic distillation of pyridine and water (feed composition: $x_{\mathrm{Py}}=0.1, x_{\mathrm{Wa}}=0.9$) using toluene as solvent, with a concept similar to {\bf b}.
	}
	\label{figure_flowsheets_lit_feeds}
\end{figure}

For every system from Table \ref{table_example_systems}, we evaluate the agent on a feed stream provided in the literature \cite{wang2018, kunnakorn2013, luyben2008, chen2015}. We show and discuss the flowsheets constructed by the agent in those situations (trained and evaluated on reward $\mathcal{C}_{\textnormal{\footnotesize literature}}$) in Figure \ref{figure_flowsheets_lit_feeds}. Similarly to the processes from the literature (not shown), the agent can separate the feed stream and the used solvent in all four cases. In all examples shown in Figure \ref{figure_flowsheets_lit_feeds}, the agent uses recycles to enable the separations and reduce waste streams. As discussed in the following section, it even chooses the continuous specifications of the units so that they only make sense in combination with those recycles. When the agent encounters different feed stream compositions than in Figure \ref{figure_flowsheets_lit_feeds} or is trained with reward $\mathcal{C}_{\textnormal{\footnotesize generic}}$, it slightly adjusts the flowsheet topology and the specifications of the unit operations (e.g., see the process in Figure \ref{figure_schrittweiser_aufbau}). 

\subsection{Recycle loops and continuous specifications}

\subsubsection{Learning from failed flowsheets}

A unique action in flowsheet synthesis is the placement of a recycle loop. While all other unit operations alter a chosen stream and potentially generate new stream(s) in the flowsheet, a recycle loop can alter all streams of the entire loop and drastically change the overall dynamics of flowsheet synthesis. In most problem instances considered, it is only possible to solve the task by placing recycles. It is important to note that recycling streams can lead to divergent flowsheet simulations within our framework. Divergent flowsheet simulations lead to failed flowsheets, and it is a known challenge how an agent can learn to avoid failed flowsheets without learning to avoid altogether placing a recycle \cite{goettl2021b, stops2023}.

We solve this problem directly in the MCTS by recursively pruning the corresponding node from the tree whenever a search simulation reaches a failed flowsheet and continuing the search from the simulation's last feasible actions. While this simple procedure leads to longer search times at the beginning of training, the search tree statistics directly reflect the pruning from which the policy is trained with a Kullback-Leibler divergence \cite{bishop2006} loss. The latter ensures that the agent can learn to assess whether a recycle might lead to a failed flowsheet, even if the action ultimately taken in the environment is not a recycle. Effectively, with only a small number of search simulations, the agent never chooses to place an infeasible recycle on the flowsheet. 

\subsubsection{Evolution of long-planned recycles}

Due to the sequential nature of the problem, the agent must set up the flowsheet topology and the continuous specifications of the units upfront to make sense when recycling. In early stages of the training process, the agent rarely places recycle loops, as they often lead to failed flowsheets. It rather focuses on learning generally good designs from the available unit operations and their continuous specifications. Only in a later stage, it liberally starts to place recycles, adjusting the unit specifications in a way which only make sense with a future recycle loop. This further implies a future recycle is a decision the agent must be aware of early on in the design. 

\begin{figure}[t!]
	\centering
	\includegraphics[width=0.95\linewidth]{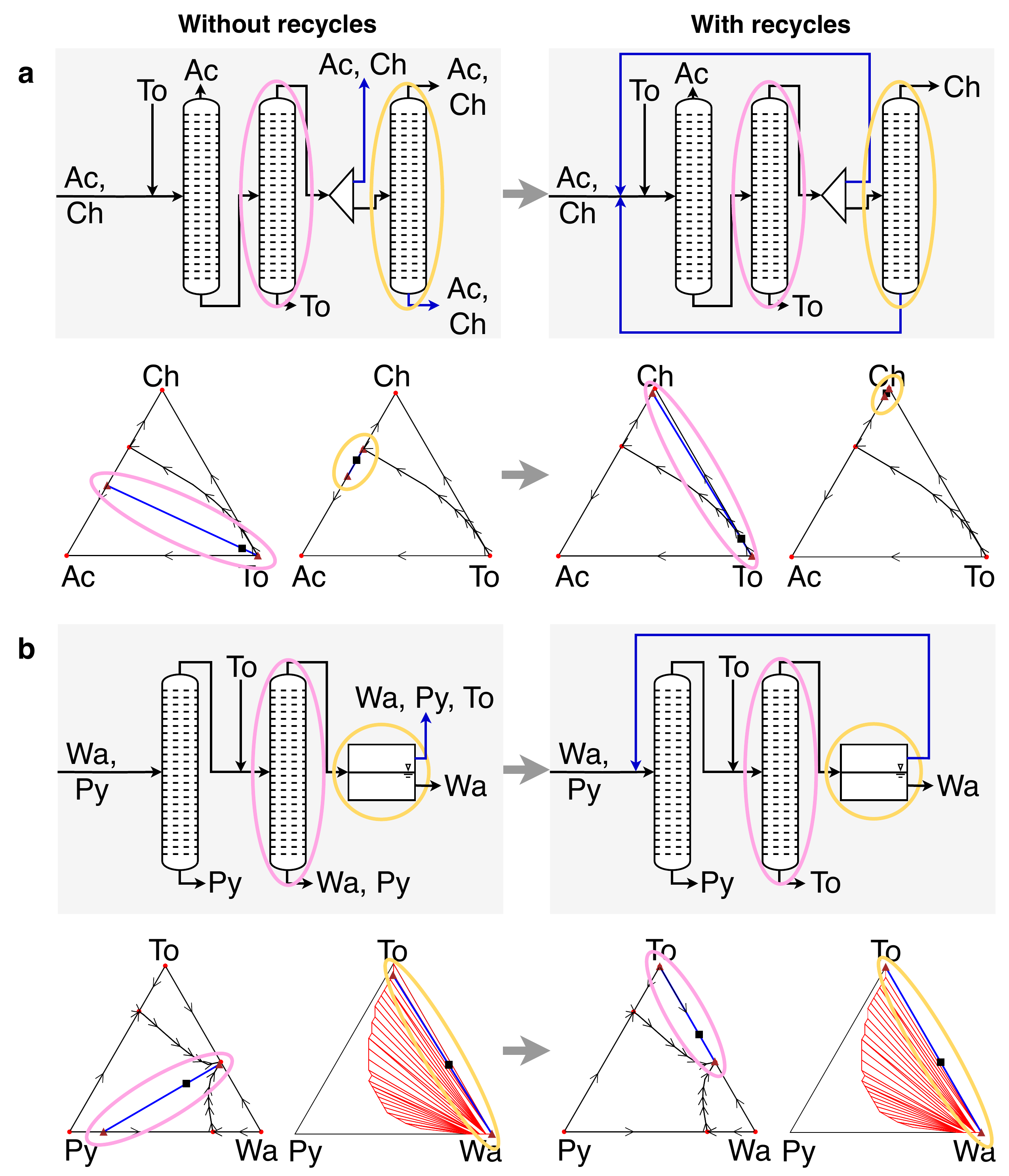}
	\caption{Examples for the implications of recycles on the compositions of the streams showing the planning capabilities of the agent trained with reward $\mathcal{C}_{\textnormal{\footnotesize generic}}$. On the left, we show the flowsheets immediately before the agent places recycles. The lower ternary diagrams visualize the two selected units (marked pink and yellow). Correspondingly, the flowsheets containing the recycles are shown on the right. Inside the ternary diagrams, $\blacksquare$ indicates the feed stream, and a brown $\blacktriangle$ indicates the output streams which are connected by a blue line. {\bf a}, Separation of acetone and chloroform using toluene as entrainer (feed composition: $x_{\mathrm{Ac}}=0.74, x_{\mathrm{Ch}}=0.26$). {\bf b}, Separation of water and pyridine using toluene as solvent (feed composition: $x_{\mathrm{Wa}}=0.04, x_{\mathrm{Py}}=0.96$).}
	\label{figure_rec_analysis}
\end{figure}

We discuss the ramifications along Figure \ref{figure_rec_analysis}, which shows two processes set up by the agent after being trained using reward $\mathcal{C}_{\textnormal{\footnotesize generic}}$. Figure \ref{figure_rec_analysis} {\bf a} shows a process for the separation of acetone and chloroform using toluene as entrainer (in contrast to Figure \ref{figure_flowsheets_lit_feeds}, where benzene serves as entrainer). Here, the agent sets four continuous specifications, i.e., the solvent ratio and the ratio of distillate to feed flowrates for all three distillation columns. On the left, the pink column separates pure toluene from a binary mixture of acetone and chloroform. From this mixture, it is impossible to separate chloroform as a pure product by distillation because of the azeotrope (the feed is on the wrong side of the azeotrope, as shown in the ternary diagram). Therefore, no pure chloroform leaves the process, and the agent seems to fail correctly setting the continuous specifications. Yet after placement of two recycle loops, the ternary diagrams show that the set ratios lead to the desired separation. 

Figure \ref{figure_rec_analysis} {\bf b} shows a process for the separation of water and pyridine using toluene as solvent. As in {\bf a}, the agent chooses a different design than in Figure \ref{figure_flowsheets_lit_feeds}). Here, the agent must set the solvent ratio and the ratio of distillate to feed flowrates for both distillation columns. As above, the continuous specifications do not yield pure product streams inside the distillation column on the left. Additionally, the decanter only separates pure water, but not pure toluene. Again, the placement of the recycle loop changes the compositions of all streams of the flowsheet (visualized in the ternary diagrams), leading to a complete separation into pure products.

\section{Discussion}

So far, RL for AFS has been explored with a rather narrow scope, training agents to deal with a single chemical system, sometimes even considering only one feed configuration. On the path to generality, we investigate for the first time whether it is possible for an agent – without prior knowledge or heuristics – to synthesize near-optimal processes for separating azeotropic mixtures that require different conceptual approaches. We decompose the discrete-continuous hybrid action space into several hierarchies which we further extend with additional hierarchy levels for factorized discretizations of continuous unit specifications. We base the agent's learning algorithm on Gumbel AlphaZero and develop a method to carefully incorporate failed flowsheets into the agent's learned policy. Furthermore, we use a sequence-to-sequence network architecture based on the lightweight MLP-Mixer to encode flowsheets. Together, this provides a general framework for tackling conceptual process design problems. The trained agent discovers azeotropic and entrainer distillation, classical techniques in chemical engineering, to successfully solve a wide range of problems from the literature. Furthermore, the learned neural network is able to capture the underlying dynamics of the systems, as it has been shown that by discarding the MCTS, the policy prediction alone can generate near-optimal flowsheets by using the much simpler beam search. The flowsheets can be generated within seconds on moderate hardware and can serve as a starting point for further process design and optimization tools that rely on a feasible process given as initialization. The methodological framework provided paves the way for further research on flowsheet simulation and the use of RL for AFS in general. Different unit operations, such as reactors and crystallizers, can be integrated into the process simulation. Their extension would be straightforward as one would only need to implement the respective unit operations into the existing framework. 

For now, the case studies are limited to processes with only up to three components, a limitation imposed by the flowsheet simulator used. To extend to more components, one would have to use more sophisticated simulation approaches for modeling vapor-liquid equilibria and distillation columns for a variable number of components.

Certainly, future research should move further toward generality, with transfer learning being a possible next step. For example, in our case study of separating feed streams into pure components, this means that the provided RL framework could be used to train an agent on a large set of chemical systems and then evaluate it on an unseen set of chemical systems to see if the agent can transfer its learning to problems it has never encountered before. 

\section{Methods}

\subsection{Environment}
\label{method_section_environment}

\subsubsection{States and actions}

The environment is a steady-state flowsheet simulation simulating a unit operation sequence given a binary system. It starts with a feed composition with molar fraction $x_1 \in (0, 1)$ of the first component and molar fraction $x_2 = 1 - x_1$ of the second component. The agent executes multiple actions on different hierarchy levels to place a unit on the flowsheet. There are three general levels, which are conceptually defined similarly to \cite{goettl2021a, goettl2021b, goettl2022b}. At level 1, the agent can terminate the flowsheet synthesis or select an open stream in the current flowsheet. If a stream is selected, the agent transitions to level 2, where it must choose a unit operation for the selected stream from a pool of available units. This unit operation is further specified (for mix/recycle stream) in level 3a or (for factorized continuous specification of distillation columns/add solvent) in level 3b, which has another sublevel 3c. The different units and their further specifications are detailed below. Once the agent has traverse all hierarchical levels, the fully specified unit operation is placed on the flowsheet, and the simulator state is updated. Throughout this work, we constrain the maximum number of units on a flowsheet to 10. Hence, if the agent does not decide to terminate earlier, the flowsheet synthesis finishes automatically after the placement of ten unit operations. For a maximum of ten units, we set the number of matrix lines to $m = 23$, which is sufficient to store all produced streams.

At each timestep $t$, the agent observes a state containing all relevant information: the flowsheet matrix $\bm M_{s_t}$ representing the process simulator's current state, the current action hierarchy level, as well as all the actions taken on previous hierarchy levels at the matrix $\bm M_{s_t}$. Note that the agent always transitions to a new state $s_{t+1}$ after each hierarchical action. In particular, the matrix $\bm M_{s_t}$ remains identical for subsequent states until all hierarchical action levels have been traversed because only then is the flowsheet matrix updated. Furthermore, we provide the agent the scalar reward it would receive at the current state if the flowsheet synthesis were terminated immediately. The construction of the flowsheet matrix from the simulator state is detailed in the Supplementary Material \ref{supp_section_flowsheet_simulation}.

After each action, the agent obtains zero reward and gets a nonzero reward (detailed below) only when the flowsheet synthesis finishes.

\subsubsection{Factorized discretization of continuous actions}

We discretize continuous actions from an interval $[\alpha, \beta] \subset \mathbb R$ with a two-step factorization: The agent chooses a discrete tuple $(m_1, m_2) \in \{1, \dots, L_1\} \times \{1, \dots, L_2\}$, where $L_1, L_2 \in \mathbb N$ are predefining the granularity of the discretization. The tuple $(m_1, m_2)$ translates to the element
\begin{equation}
	\alpha + (m_1 - 1) \cdot\frac{\beta - \alpha}{L_1} + (m_2 - 1) \cdot \frac{\beta - \alpha}{L_1L_2} \in [\alpha, \beta].
\end{equation}

In particular, the agent chooses $(m_1, m_2)$ sequentially by picking $m_1$ at action hierarchy level 3b and then $m_2$ at level 3c. This factorization is appealing as the agent can first make a coarse-grained decision via $m_1$ and then refine it with $m_2$. Thus, the agent learns more quickly which interval ranges are suitable instead of choosing from a single discrete distribution of size $L_1L_2$. In this work, we set $L_1 = L_2 = 7$ everywhere, effectively dividing the interval $[\alpha, \beta]$ into 49 evenly spaced actions.

\subsubsection{Unit Operations} \label{methods:unit_ops}

The following unit operations and specifications are available as actions to the agent.
\begin{enumerate}
	\item [I)] \bfseries Distillation column\mdseries\\
	When a distillation column is chosen at level 2, it is further specified at level 3b (with sublevel 3c) by setting the continuous ratio of distillate flowrate $\dot{n}^{\textnormal{\footnotesize D}}$ to feed flowrate $\dot{n}^{\textnormal{\footnotesize F}}$:
	\begin{equation}
	\frac{\dot{n}^{\textnormal{\footnotesize D}}}{\dot{n}^{\textnormal{\footnotesize F}}}\in[0, 1].
	\end{equation}
	In this work, we specify the pressure in the column to 1 bar, cf. Table~\ref{table_example_systems}. The height and reflux ratio are assumed infinite, giving the maximum performance. In this case, the single value $\frac{\dot{n}^{\textnormal{\footnotesize D}}}{\dot{n}^{\textnormal{\footnotesize F}}}$ fully specifies the column in simulation.
	\item [II)] \bfseries Decanter\mdseries\\
	No further specification is required when a decanter is chosen at hierarchy level 2, as we assume constant temperature and pressure. The decanter splits the feed stream according to the underlying liquid phase equilibrium. We consider only systems with binary liquid phase splits. If there is no liquid phase split for the given feed composition, the decanter splits the feed flowrate with a ratio of 1:1 into two streams of equal composition.
	\item [III)] \bfseries Mix stream\mdseries\\
	In this case, the stream chosen at level 1 is mixed with another open stream picked at level 3a. Therefore, the specification is a purely discrete decision.
	\item [IV)] \bfseries Recycle stream\mdseries\\
	In this case, the stream chosen at level 1 is recycled back. Similarly to 'Mix stream', the (discrete) destination is specified at level 3a.
	\item [V)] \bfseries Add solvent\mdseries\\
	Table \ref{table_example_systems} shows available solvents for the considered chemical example systems. To constrain the computational complexity during the flowsheet simulation, we limit the total number of components to 3. Thus, the agent can select this unit operation once per flowsheet synthesis process. The solvent is added to the stream chosen at level 1. At level 3b and 3c, the agent must specify the continuous ratio of solvent flowrate $\dot{n}^{\textnormal{\footnotesize S}}$ to the flowrate of the chosen stream $\dot{n}^{\textnormal{\footnotesize F}}$:
	\begin{equation}
	\frac{\dot{n}^{\textnormal{\footnotesize S}}}{\dot{n}^{\textnormal{\footnotesize F}}}\in[0, 10].
	\end{equation}
	The ratio is limited to 10 in our framework, giving ample space to solve the considered problems. In general, however, larger ratios could be allowed.
\end{enumerate}
Short-cut models simulate the available unit operations to provide a fast and robust environment for the RL agent. The $\infty$/$\infty$-approach \cite{ryll2009, bekiaris1996, ryll2012b} and the convex envelope method \cite{ryll2009, ryll2012a, goettl2023} are employed to simulate distillation columns and decanters, respectively. We provide details within the Supplementary Material \ref{supp_section_flowsheet_simulation}.

\subsubsection{Reward \texorpdfstring{$\mathcal{C}_{\textnormal{\footnotesize literature}}$}{(literature)}}

The reward $\mathcal{C}_{\textnormal{\footnotesize literature}}$ evaluates a given flowsheet by calculation of the net present value (NPV) in a similar way as presented in \cite{goettl2021b}, i.e.,
\begin{equation}
\mathcal{C}_{\textnormal{\footnotesize literature}} = -\sum_{u\in U}C_u + 10\textnormal{a}\big(-\sum_{u\in U}C_{\textnormal{\footnotesize op}, u}+\sum_{o\in O}c_o\big).
\end{equation}
All parameters and prices for $\mathcal{C}_{\textnormal{\footnotesize literature}}$ are provided within the Supplementary Material \ref{supp_section_c_tables}. $U$ is the set of all unit operations used in the process. For every unit operation $u\in U$, $C_u$ describes the total investment costs and $C_{\textnormal{\footnotesize op}, u}$ the annual operating costs, respectively. We denote by $O$ the set of all streams leaving the process. For every $o\in O$ leaving, $c_o$ describes the annual operational cash flow. 

The investment costs are scaled depending on the feed mass flowrate of the unit operations according to the power rule \cite{towler2022}:
\begin{equation}
C_u = C_{0, u}\cdot\Big(\frac{\dot{m}_u}{\dot{m}_{0, u}}\Big)^{0.6}.
\end{equation}
The operating costs comprise only steam costs for distillation columns and costs for added solvents. Steam costs are calculated via
\begin{equation}
C_{\textnormal{\footnotesize op, distillation column}} = p_\textnormal{\footnotesize steam}\cdot\frac{\dot{Q}_\textnormal{\footnotesize Reboiler}}{\Delta h^{(m)}_{\textnormal{\footnotesize Wa}, v}}\cdot 8000\frac{\textnormal{\footnotesize h}}{\textnormal{\footnotesize a}},
\end{equation}
where $\Delta h^{(m)}_{\textnormal{\footnotesize Wa}, v}$ is the enthalpy of evaporation of water, and $p_\textnormal{\footnotesize steam}$ is the specific cost of steam. The reboiler duty $\dot{Q}_\textnormal{\footnotesize Reboiler}$ of a distillation column is estimated from the simple assumption that the distillate has to be evaporated twice:
\begin{equation}
\dot{Q}_\textnormal{\footnotesize Reboiler} = 2\cdot\sum_{i}\dot{m}_{i, \textnormal{\footnotesize distillate}}\cdot\Delta h^{(m)}_{i, v}.
\end{equation}
For an arbitrary component $i$, $\Delta h^{(m)}_{i, v}$ is estimated by calculating the energy required to heat component $i$ at ambient conditions (298.15 K, 1 bar), in liquid form, to its boiling point and adding the heat of evaporation. Heat capacities are taken from \cite{nist2023, ddbst2023}. The heat of evaporation is computed using the Antoine equation (parameters provided within \cite{aspen2015}) and the Clausius-Clapeyron equation \cite{krey2007}. When adding a solvent, the amount has to be paid for, i.e.,
\begin{equation}
C_{\textnormal{\footnotesize op, add solvent}} = p_\textnormal{\footnotesize solvent}\cdot\dot{m}_{\textnormal{\footnotesize solvent}}\cdot 8000\frac{\textnormal{\footnotesize h}}{\textnormal{\footnotesize a}},
\end{equation}
with price $p_\textnormal{\footnotesize solvent}$.
Consider a stream $o\in O$ leaving the process. We want to assign a positive value for $c_o$ if it is (almost) a pure stream. Therefore, if the mass fraction for some component $i$ is greater than 0.99, we consider $o$ to be pure and calculate $c_o$ as
\begin{equation}
c_o = p_i\cdot\dot{m}_o\cdot 8000\frac{\textnormal{\footnotesize h}}{\textnormal{\footnotesize a}},
\end{equation}
where $p_i$ is the price of the respective component and $\dot{m}_o$ is the mass flow rate of stream $o$. If the defined specification (some component $i$ with a mass fraction greater than 0.99) is not fulfilled, $c_o$ is set to $0$. Note that a solvent can be added practically for free if it is entirely separated after using it for separation purposes.

\subsubsection{Reward \texorpdfstring{$\mathcal{C}_{\textnormal{\footnotesize generic}}$}{(generic)}}

As mentioned, $\mathcal{C}_{\textnormal{\footnotesize generic}}$ is a generic reward function, which is particularly useful for conceptual flowsheet synthesis. $\mathcal{C}_{\textnormal{\footnotesize generic}}$ assigns a large value (without monetary unit) to processes that separate the feed stream into pure components and does not take into account an economical analysis of the process. Secondary objectives are, for example, to use as few unit operations as possible and to obtain the added solvent as pure stream after usage. We set
\begin{equation}
\mathcal{C}_{\textnormal{\footnotesize generic}} = -\sum_{u\in U}C_u-C_{\textnormal{\footnotesize op, add solvent}}+\sum_{o\in O}c_o,
\end{equation}
where we assign constant costs for every unit operation. Contrary to $\mathcal{C}_{\textnormal{\footnotesize literature}}$, we omit scaling of unit costs with size and steam costs for distillation columns. When a solvent is added, the cost is computed as
\begin{equation}
C_{\textnormal{\footnotesize op, add solvent}}=p_{\textnormal{\footnotesize solvent}}\cdot\dot{n}_{\textnormal{\footnotesize total added solvent}}.
\end{equation}
Similarly, as before, $C_{\textnormal{\footnotesize op, add solvent}}$ can be neglected by separating all used solvents into (almost) pure streams. 

For a product stream $o\in O$, we calculate the gain $c_o$ if it is (almost) a pure stream, which we define to be a molar fraction greater than 0.99 for an arbitrary component $i$:
\begin{equation}
c_o = p_i\cdot\dot{n}_o.
\end{equation}
All parameters for $\mathcal{C}_{\textnormal{\footnotesize generic}}$ are provided in the Supplementary Material \ref{supp_section_c_tables}.

\subsubsection{Performance Ratio \texorpdfstring{$\mathcal{R}$}{R}}

We define a performance ratio $\mathcal{R}\in[0, 1]$ for the flowsheets examined in this work. Consider a flowsheet with feed stream $(\dot{n}_1^{\textnormal{\footnotesize F}}, \dot{n}_2^{\textnormal{\footnotesize F}}, 0)$. Let $\dot{n}_3^{\textnormal{\footnotesize S}}$ be the accumulated amount of a solvent, which was added to the process (note that $\dot{n}_3^{\textnormal{\footnotesize S}}$ can also be equal to $0$). Let $O_{\textnormal{\footnotesize spec}}=\{o_1, \dots, o_K\}$ be the set of leaving streams that meet the purity specification (depending on $\mathcal{C}_{\textnormal{\footnotesize literature}}$ or $\mathcal{C}_{\textnormal{\footnotesize generic}}$). This means that for all $o_i=(\dot{n}_{1, i}, \dot{n}_{2, i}, \dot{n}_{3, i})$ with $i=1,\dots,K$ it holds that there exists exactly one $j\in\{1,2,3\}$ so that $=x^{(m)}_{j, i}>0.99$ (for $\mathcal{C}_{\textnormal{\footnotesize literature}}$) or $=x_{j, i}>0.99$ (for $\mathcal{C}_{\textnormal{\footnotesize generic}}$). We define $\mathcal{R}$ as:
\begin{equation}
\mathcal{R}=\frac{1}{\dot{n}_1^{\textnormal{\footnotesize F}}+\dot{n}_2^{\textnormal{\footnotesize F}}+\dot{n}_3^{\textnormal{\footnotesize S}}}\cdot\Big(\sum_{i=1}^{K}\dot{n}_{1, i}+\dot{n}_{2, i}+\dot{n}_{3, i}\Big).
\end{equation}
$\mathcal{R}$ measures how much of the input of a process is separated into pure streams ('pure stream' is defined by the respective specification). 

\subsection{Learning algorithm}
\label{method_learning_algorithm} 

We train the agent with Gumbel AlphaZero \cite{danihelka2022}, an algorithmic redesign of AlphaZero \cite{silver2018}, where a neural network guides an MCTS. 
The neural network $g_\theta$ is parameterized by $\theta$ and takes as input a state $s_t$ and outputs a tuple $g_\theta(s_t) = (\pi(\cdot | s_t), v(s_t))$, where $\pi(\cdot | s_t)$ is the policy at the state $s_t$, i.e., a probability distribution over actions at the hierarchical level corresponding to $s_t$, and where $v(s_t)$ is a prediction of the expected reward the agent obtains when continuing the synthesis from $s_t$. Multiple actor processes keep frozen network parameters and let the agent run episodes from random initial configurations in parallel, where at each encountered state $s_t$, an MCTS is performed. After the tree search at $s_t$, the search statistics yield an improved policy $\hat \pi(\cdot | s_t)$ and action $a_t$ at the current hierarchy of $s_t$. The agent takes action $a_t$ in the environment, transitions to state $s_{t+1}$, and the search tree is shifted to the subtree under $a_t$ to use for the search at $s_{t+1}$. When the episode terminates, the agent obtains the reward $r \in \mathbb R$ and sends it to a network training process with the improved policies at all intermediate states of the trajectory. The training process uses the improved policy $\hat \pi(\cdot | s_t)$ to update the network's policy $\pi(\cdot | s_t)$ using a Kullback-Leiber divergence loss $\mathrm{KL}(\hat \pi || \pi)$, and the value prediction $v(s_t)$ is updated via the squared error $(v(s_t) - r)^2$. The training process periodically disseminates the updated network parameters $\theta$ to the actor processes for further episodes.

\subsubsection{Action selection in Gumbel AlphaZero}

In the search tree, nodes represent states, and edges represent actions. As in AlphaZero, a single search simulation from a root node $s$ traverses the tree until reaching a leaf node $s_L$, which is evaluated by the neural network $g_\theta(s_L) = (\pi(\cdot | s_L), v(s_L))$. Then, the leaf node is expanded using $\pi(\cdot | s_L)$, and the predicted value $v(s_L)$ is recursively backed up the trajectory. In particular, as in AlphaZero, we store for all edges the search statistics $N(s, a)$ and $Q(s, a)$, where $N(s, a)$ denotes the visit count and $Q(s, a)$ is the estimated action value (i.e., accumulated backed up values divided by $N(s,a)$).

Given a root node $s$, we sample a maximum of $z=16$ actions without replacement from the predicted policy $\pi(\cdot | s)$ using the Gumbel-Top-k trick \cite{yellott1977relationship,vieira2014gumbel}. We denote by $\text{logit}_\pi(a)$ the (unnormalized) log-probability of action $a$ and by $G(a)$ its sampled Gumbel noise. Using a Sequential Halving \cite{karnin2013almost} procedure, a predefined budget of simulations is evenly distributed between the sampled actions, and multiple search simulations start from each sampled action. After each Sequential Halving level, the considered root actions are pruned to the top $z \gets \frac{z}{2}$ actions according to their scores 
\begin{equation} \label{methods:eq:root_scoring}
G(a) + \text{logit}_\pi(a) + \sigma(Q(s,a)),
\end{equation} where $\sigma$ is the monotonically increasing linear map
\begin{equation}
	\sigma(Q(s, a)) = (50 + \max_b(N(s,b))) Q(s,a),
\end{equation}
matching the choice in \cite{danihelka2022}. When the search budget is exhausted, the agent chooses an action from the remaining unpruned actions with maximum scoring (\ref{methods:eq:root_scoring}). In this work, we grant a budget of 200 simulations at a root node. With 16 sampled actions, this amounts to four Sequential Halving levels with 16, 8, 4, and 2 remaining root actions, where each remaining action gets 3, 6, 12, and 28 simulations. 

After the MCTS, the improved policy $\hat \pi(\cdot | s)$, which serves as a training target, is constructed in a two-step process. First, a 'completed Q-value' $\hat Q(s, a)$ is defined for all actions $a$ by setting
\begin{equation}
	\hat Q(s, a) := 
	\begin{cases}
  		Q(s, a) & \text{if }N(s, a) > 0, \\
  		\hat v(s) & \text{else}.
  \end{cases}
\end{equation}
Here, $\hat v(s)$ is defined as the interpolation
\begin{equation}
	\hat v(s) = \frac{1}{1 + \sum_b N(b)} \left(v(s) + \frac{\sum_b N(b)}{\sum_{b, \text{ s.t. }N(b) > 0}\pi(b|s)} \sum_{a, \text{ s.t. }N(a) > 0}{\pi(a|s) Q(s, a)} \right).
\end{equation}
Then, $\hat \pi(\cdot | s)$ is constructed by setting for all actions $a$
\begin{equation}\label{methods:eq:visit_count_dist}
	\hat \pi(a | s) := \text{softmax}(\text{logit}_{\pi}(a) + \sigma(\hat Q(s, a)).
\end{equation}
Informally, the improved policy $\hat \pi$ increases the logit of an action where the search tells us that it leads, on average, to higher returns than expected. It decreases the logit otherwise while giving zero advantage to unvisited actions. 

During the search, at a non-root node $\tilde s$, we can compute $\hat \pi(\cdot | \tilde s)$ according to (\ref{methods:eq:visit_count_dist}) and select an action deterministically from
\begin{equation} \label{method:eq:intree}
	\argmin_a \sum_{b}\left(\hat \pi(b | \tilde s) - \frac{N(b) + {\bf 1}_{\{a\}}(b)}{1 + \sum_c N(c)}\right)^2,
\end{equation}
where ${\bf 1}_{\{a\}}(b) = 1$ if $a = b$, and zero otherwise. Intuitively, this deterministic action selection chooses the action that shifts the visit count distribution closer to $\hat \pi$.

As in \cite{danihelka2022}, whenever computing (\ref{methods:eq:root_scoring}) or (\ref{methods:eq:visit_count_dist}), we normalize the Q-values with a min-max normalization according to all Q-values encountered in the search tree so far.

\subsubsection{Infeasible actions, flowsheets, and tree pruning}

The substantial advantage of using the improved policy $\hat \pi$ as a target for the network, trained with Kullback-Leiber divergence, is that $\hat \pi$ incorporates rich information from the search, as opposed to classically training the network to predict the single action which the agent takes in the environment. We leverage this effect and mask infeasible actions everywhere they are encountered in the tree by setting their corresponding logit to $- \infty$ before computing $\hat \pi$. Through this, the network learns to reduce the predicted probability for infeasible actions, better capturing the system's dynamics. We distinguish between two types of infeasible actions. First, infeasible actions arise directly from the definition of the environment. These are all closed streams in action level 1, 'add solvent' in level 2 if a solvent is already present, all closed streams in level 3a when mixing streams (as an open stream can only be mixed with another open stream), and all open streams in level 3a when recycling a stream (as an open stream can only be recycled to a closed one). The second type of infeasible action arises when recycling a stream and choosing a destination at level 3a such that the simulator does not converge. These actions are unknown upfront and must be tried during the tree search. We devise a recursive tree pruning technique to address the second type of infeasible actions, which we call divergent actions in the following for simplicity: Whenever a search simulation encounters a divergent action at a state $s$ on level 3a, we set its logit to $-\infty$ and the action selection at $s$ according to (\ref{method:eq:intree}) is repeated after recomputing the improved policy $\hat \pi$. If all actions at $s$ are divergent, the node $s$ and its subtree are pruned, and the action from the parent of $s$ leading to $s$ is set as infeasible, repeating the simulation from the parent of $s$. We repeat this process recursively. Note that the termination action is always allowed at hierarchical action level 1. Hence, coupled with Sequential Halving (where in the last Sequential Halving step, the remaining simulations are distributed only between two actions), it is unlikely that after the search, the agent decides to execute an action in the environment that eventually leads to a divergent action. In practice, we never observe the agent executing such an action, so no experience is discarded. Furthermore, as the improved policy directly reflects divergent actions, their infeasibility is recursively accounted for in $\hat \pi$ at the root state, so the network is trained to give a lower probability to these actions. Thus, as training progresses, the number of tree prunings becomes smaller and smaller without the agent learning to avoid recycling streams altogether.

\subsubsection{Training cycle}

We train the agent for 50k episodes, where 50 parallel actor processes generate trajectories with frozen network parameters. After each episode, the actor sends the trajectory of states, actions, final reward, and improved policies at all states to a replay buffer process. For each action hierarchy level, a separate network training process samples uniformly random batches of states of size 128 with replacement from the replay buffer and performs an optimizer step with respect to the value and policy on the given level. It ignores a hierarchy level if the number of trajectories containing at least one action from that level does not exceed a predefined threshold. This avoids overfitting for later levels at the beginning of training. We experienced that a threshold of 50 is generally enough. 
One optimizer step for all hierarchy levels (including skipped ones) constitutes one training step. We constrain the agent to a ratio of approx. 3 training steps to 1 episode throughout training. After every 100 training steps, the updated network parameters are distributed to the actor process to refresh their frozen network copies. Furthermore, after every 7.5k steps, the performance of the current agent is evaluated on a pre-generated random but fixed validation set of 200 initial states. The best-performing agent on the validation set is eventually used for testing after training.
We train the agent with an AMD EPYC 7543 32-core processor and two NVIDIA RTX A5000, each with 24GB of memory. Training for 50k episodes from zero knowledge takes about 1.5 days. 

\subsection{Neural network}

The neural network takes as input a state $s$ and outputs logits for all action hierarchy levels and the predicted value $v(s)$ simultaneously. It consists of two identically structured (cf. Fig.~\ref{figure_method}b) MLP-Mixer \cite{tolstikhin2021mlp} networks with separate weights (one for the policies, one for the value) for representing the streams, followed by a total of four separate policy heads for the action hierarchy levels (where a single head predicts level 3b and c together) and a single value head. The following describes the input and its way through the network. In this work, the latent space $\mathbb R^d$ is of dimension $d = 128$ everywhere.

\subsubsection{State representation} 

Consider a state $s$ at which an action must be chosen with (by abuse of notation) hierarchy level $l \in \{1,2,3\mathrm a,3 \mathrm b,3 \mathrm c\}$. Again, denote by $\bm M_s \in \mathbb R^{m \times n}$ the flowsheet matrix corresponding to $s$, constructed as described in the supplementary material \ref{supp_section_flowsheet_simulation}. Let $r \in \mathbb R$ be the reward the agent would obtain if the flowsheet synthesis were terminated with $\bm M_s$. Let $\bm w_1, \dots, \bm w_m \in \mathbb R^n$ be the (potentially zero, if the matrix is not full) lines of $\bm M_s$ corresponding to the streams. If $l \in \{2,3\mathrm a,3 \mathrm b,3 \mathrm c\}$, then an open stream has been chosen in a previous state, and we denote its index by $i_{\text{stream}} \in \{1, \dots, m\}$. If $l = 1$, we define $i_{\text{stream}} := 0$. Furthermore, denote by $\bm e_{\text{units}} \in \mathbb R^{\text{\#units}}$ a one-hot-encoded vector that indicates the chosen unit of level 2 (and a zero-vector if $l \in \{1,2\}$). Similarly, we let $\bm e_{\text{cont}} \in \mathbb R^{L_1}$ be the one-hot-encoded vector indicating which variable was chosen in the discretization of level 3b (i.e., this is only nonzero if $l$ is 3c).

\subsubsection{Torso network}

The torso network consists of two MLP-Mixer \cite{tolstikhin2021mlp} networks $\text{MLP-Mixer}^v$ and $\text{MLP-Mixer}^\pi$, which have identical architectures but have different network parameters. Let $I := \{1, \dots, m\}$. We embed each stream $\bm w_1, \dots, \bm w_m$ into $\mathbb R^d$ with two learnable affine maps $H^v, H^\pi \colon \mathbb R^n \to \mathbb R^d$ and transform the embedded sequences via
\begin{equation}
	\left(\tilde{\bm w}_i^v\right)_{i \in I} := \text{MLPMixer}^v\left((H^v(\bm w_i))_{i \in I}\right) \in \mathbb R^{m \times d}
\end{equation}
and analogously
\begin{equation}
	\left(\tilde{\bm w}_i^\pi\right)_{i \in I} := \text{MLPMixer}^\pi\left((H^\pi(\bm w_i))_{i \in I}\right) \in \mathbb R^{m \times d}.
\end{equation}

Both MLP-Mixers consist of 5 mixer blocks whose design follows the original architecture \cite{tolstikhin2021mlp}. In each mixer block, we use layer normalization, a hidden dimension of 512 in the feature mixing MLP, and a hidden dimension of 2$m$ in the token mixing MLP. 
The sequence $\left(\tilde{\bm w}_i^\pi\right)_{i \in I}$ is used as input for all policy heads and $\left(\tilde{\bm w}_i^v\right)_{i \in I}$ is used for the value head.

\subsubsection{Policy heads}

We index the four policy heads corresponding to each hierarchy level by $j$. The heads are structurally the same, taking in the sequence $\left(\tilde{\bm w}_i^\pi\right)_{i \in I}$, the chosen stream index $i_{\text{stream}}$ and an optional head-specific vector $\bm e^j$, detailed below. A head applies a final single MLP-Mixer block 
\begin{equation}
\text{MixerBlock}^j(\left(\tilde{\bm w}_i^\pi\right)_{i \in I}) =: \left(\tilde{\bm w}_i^{\pi, j}\right)_{i \in I}
\end{equation}
and computes a linear (with learnable coefficients) combination $\bm h^j \in \mathbb R^d$ of the resulting sequence. We concatenate $\bm h^j$, $\tilde{\bm w}_{i_{\text{stream}}}^{\pi, j}$ (if $i_{\text{stream}} > 0$) and $\bm e^j$ (if present, see below) and apply the resulting vector to a final MLP outputting logits for the policy on the given level.
The specifics for the four policy heads are as follows:
\begin{itemize}
	\item Level 1 (choose open stream): Output dimension $m$, no head-specific vector. 
	\item Level 2 (choose unit): Output dimension $\#units$, no head-specific vector.
	\item Level 3a (choose second stream): Output dimension $m$, head-specific vector $\bm e_{\text{units}}$
	\item Level 3b and c (choose continuous spec): Output dimension $L_1 + L_2$, the head-specific vector is a concatenation $[\bm e_{\text{units}};\bm e_{\text{cont}}]$.
\end{itemize}

\subsubsection{Value head}

The value head operates on the sequence $\left(\tilde{\bm w}_i^v\right)_{i \in I}$ and is structurally similar to the policy head for Level 3b and c (with output dimension 1), but additionally concatenates the reward-if-terminated $r$ to the head-specific vector. 

\subsection{Search at inference time}

We test the agent with two search settings, the Gumbel AlphaZero MCTS as it was trained with and generating trajectories using only the policy with beam search. 

\subsubsection{MCTS} During inference, we do not sample actions without replacement at a root node but take the top 16 actions with the highest logits (equivalent to the Gumbel-Top-k trick with zero Gumbels). Consequently, we also set the Gumbel noise $G(a)$ to zero when computing (\ref{methods:eq:root_scoring}). As the in-tree action selection is, by design, already deterministic, the agent becomes fully deterministic at inference time.

\subsubsection{Beam search} We briefly describe how to perform beam search with beam width $k \in \mathbb N$ and a policy network in general to obtain a set of high-probability sequences. For any state $s$, denote by $\mathcal A(s)$ the set of feasible actions from $s$ and by $s \circ a$ the state obtained when executing $a \in \mathcal A(s)$. Let $s_0$ be an initial state, and set $\textsc{beam} := \{s_0\}$. A beam search recursively applies the following two steps until all states in $\textsc{beam}$ are terminal:
\begin{enumerate}
	\item [I)] Expand beam
	\begin{equation}
		\textsc{beam} \gets \{s \circ a \mid s \in \textsc{beam} \text{ not terminal, } a \in \mathcal A(s) \}.
	\end{equation}
	\item [II)] Prune beam
	\begin{align}
	\textsc{beam} \gets & \text{top $k$ states $s = s_0 \circ a_1 \circ \cdots \circ a_t \in \textsc{beam}$ according to} \\ & \sum_{j = 1}^t \log \pi\left(a_j \mid s_0 \circ \cdots \circ a_{j-1}\right).
	\end{align}
\end{enumerate}

This procedure is a standard beam search over a factorized sequence model. We discard the corresponding trajectory whenever a divergent action stays in the beam after pruning. Whenever a state is terminal, we remove the trajectory from the beam and store it in a separate terminal set before continuing the beam search. The results reported correspond to the best trajectory in the terminal set.

\section{Code availability} 
Python codes based on PyTorch for training and evaluating the discussed algorithm are available at \url{https://github.com/grimmlab/drl4procsyn}.

\section*{Acknowledgement}
This work was funded by the Deutsche Forschungsgemeinschaft (DFG, German Research Foundation)
– 466387255 – within the Priority Programme ”SPP 2331: Machine Learning in Chemical
Engineering”.

\printbibliography


\clearpage

\begin{center}
	\textbf{\large Appendix}
\end{center}
\appendix

\section{Simulation framework}
\label{supp_section_flowsheet_simulation}

\subsection{Unit operations}

\subsubsection{General remarks}
The agent interacts with a steady-state flowsheet simulation, which simulates the chosen unit operations sequentially. The unit operations are based on short-cut models, which allow a quick and robust evaluation. Apart from recycle loops, all models always converge.

The phase equilibria are modeled using extended Raoult's law (vapor-liquid) or $g^E$ minimization (liquid-liquid). The non-ideality in the liquid phase is modeled using the Non-Random-Two-Liquid (NRTL) $g^E$-model \cite{prausnitz1999}. The NRTL parameters for all considered binary subsystems are taken from Aspen Plus \cite{aspen2015}. Within all of the considered units, we assume constant conditions as described in Table \ref{table_example_systems}.

\subsubsection{Distillation column}
\begin{figure}
	\includegraphics[width=0.9\linewidth]{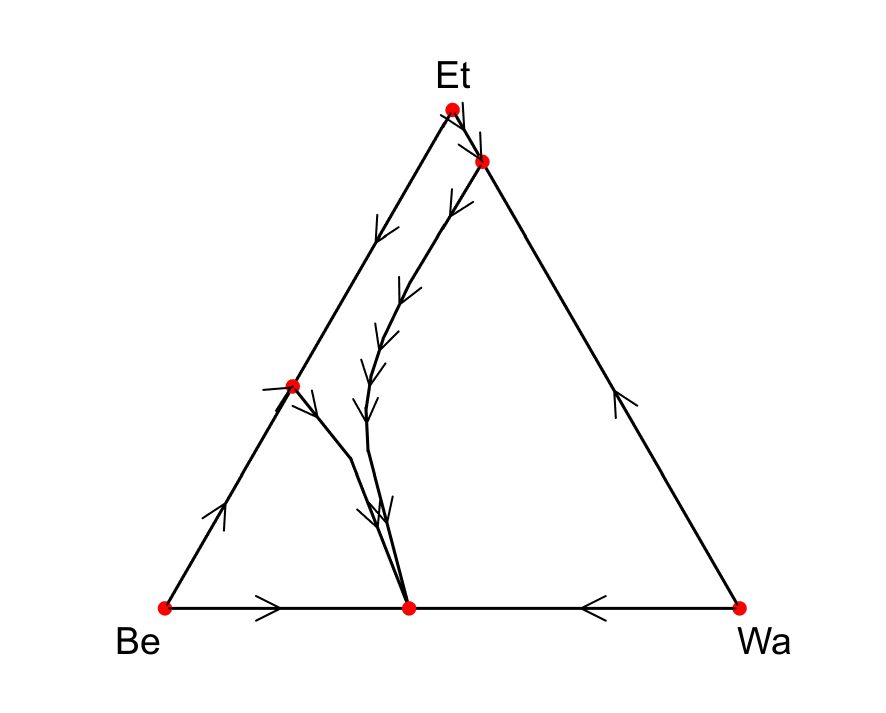}
	\caption{Linearized vapor-liquid equilibrium for the ternary system: Be, Et, Wa (at 1 bar). Singular points are marked red. The black lines separate the distillation regions (arrow direction from high-boiling towards low-boiling singular points).}
	\label{supp_figure_vle}
\end{figure}

This work uses piecewise linearized representations for vapor-liquid equilibria and distillation lines. This form of representation allows fast and robust modeling of a distillation column using the $\infty$/$\infty$-approach \cite{ryll2009, bekiaris1996, ryll2012b}, assuming an infinite number of stages and total reflux. These assumptions model a thermodynamic limiting case, which displays similar behavior as actual distillation columns. For a general overview regarding phase diagrams used to analyze distillation processes, we refer to \cite{petlyuk2004}. For a detailed description of the $\infty$/$\infty$-approach, we refer to \cite{ryll2009, ryll2012b} and outline the basic concepts in the following for the ternary system shown in Figure \ref{supp_figure_vle}.

Figure \ref{supp_figure_vle} displays the distillation boundaries for the ternary system benzene -- ethanol -- water at 1 bar. The red points indicate singular points, i.e., pure components and azeotropes. The distillation boundaries separate the distillation regions, and the arrows display piecewise linearized distillation lines, which means they are directed from a high-boiling towards a low-boiling singular point. For a detailed explanation of the construction process of this diagram, we refer to \cite{ryll2009}.

Assuming an infinite number of stages, total reflux, and constant pressure allows specification of a distillation column with one parameter: the ratio of distillate to feed flowrates $\dot{n}^{\textnormal{\footnotesize D}}/\dot{n}^{\textnormal{\footnotesize F}}$. Given a feed stream composition, distillate and bottom product are determined using the following rules:
\begin{enumerate}
	\item [I)] Feed, distillate, and bottom product are located on one straight line, which satisfies the lever arm rule and the specified value of $\dot{n}^{\textnormal{\footnotesize D}}/\dot{n}^{\textnormal{\footnotesize F}}$. 
	\item [II)] Distillate and bottom product are on the same distillation boundary. Note that this implies that the distillate and bottom product are in the same distillation region.
	\item [III)] One of the following three cases applies: the distillate is the local low-boiler (i.e., the low-boiler of this distillation region); the bottom product is the local high-boiler (i.e., the high-boiler of this distillation region); the distillation line, where distillate and bottom product are located on, passes a saddle point (i.e., a singular point, which is neither the low- or high-boiler of this distillation region).
\end{enumerate}
Using this concept, distillation columns can be modeled quickly and robustly, even for chemical systems, which display complicated, azeotropic behavior.

\subsubsection{Decanter}
\begin{figure}
	\includegraphics[width=0.9\linewidth]{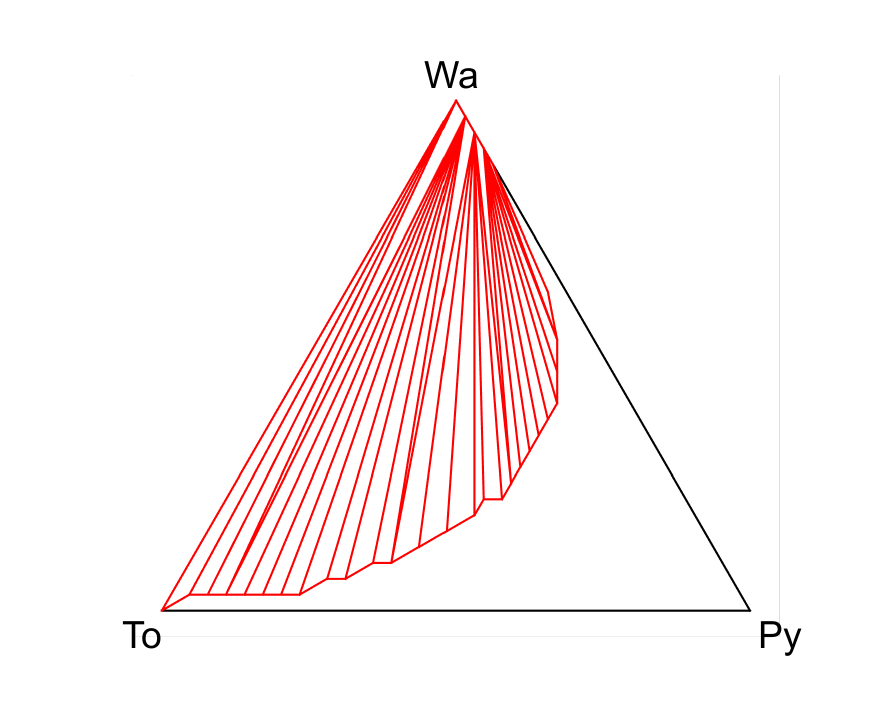}
	\caption{Liquid-liquid equilibrium constructed by the convex envelope method for a ternary system consisting of pyridine, toluene, and water (at 1 bar, 338.15 K). The red simplices model a split into two liquid phases.}
	\label{supp_figure_lle}
\end{figure}

A discretized, linearized model of the whole composition space represents the liquid-liquid equilibria. Using this representation, decanters can be simulated robustly. The construction of this representation is based on the convex envelope method, and we refer to \cite{ryll2009, ryll2012a, goettl2023} for a detailed description. Figure \ref{supp_figure_lle} visualizes the liquid-liquid equilibrium constructed by the convex envelope method for a ternary example system consisting of pyridine, toluene, and water (at 1 bar, 338.15 K). The red simplices model the two-phase region. 

\subsubsection{Recycle stream}
We use tear streams to simulate recycle loops \cite{biegler1997}. The underlying fixed-point problem is transformed into a root-finding problem and solved by the usage of the function \texttt{fsolve} provided within the Python package \texttt{scipy} \cite{virtanen2020}. We limit the flowrate of each recycle loop to a maximum of 25 times the flowrate of the feed stream of the current process.

\subsection{State representation}
\begin{figure}
	\includegraphics[width=0.9\linewidth]{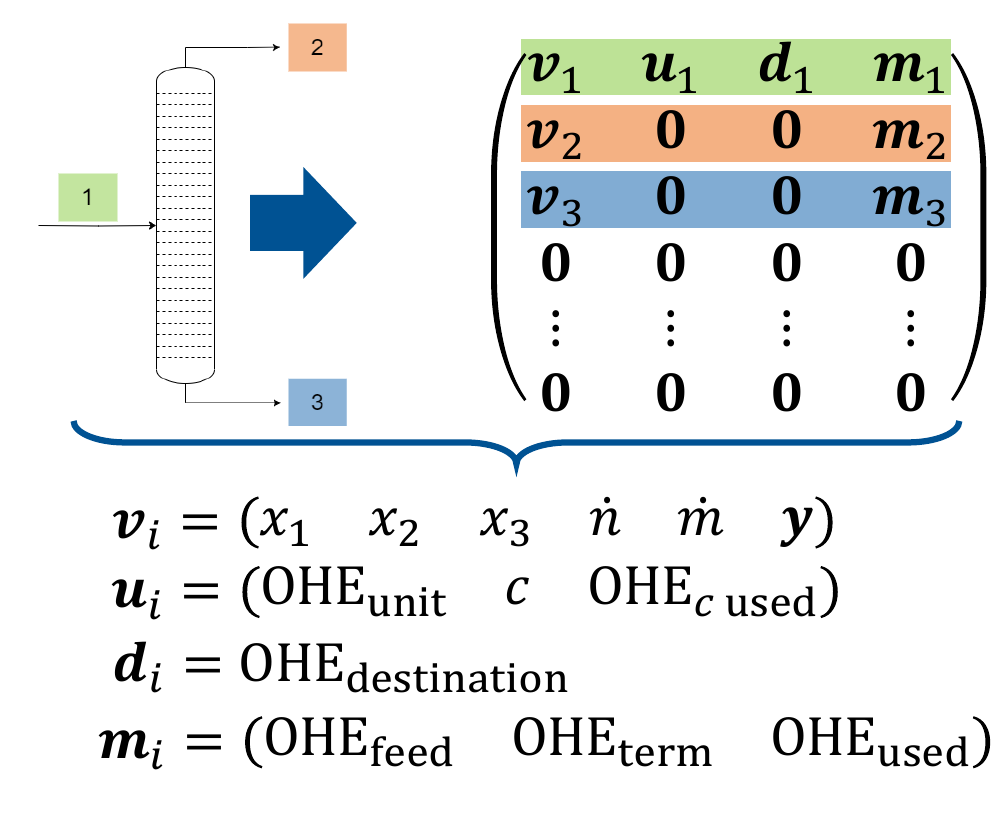}
	\caption{Matrix representation of a flowsheet. As the colors indicate, every stream corresponds to a line in the matrix.}
	\label{supp_figure_matrix}
\end{figure}

The flowsheet is represented as a matrix, which contains all information on the current process’s state. It is constructed similarly to the description in \cite{goettl2022a, goettl2021b} and augmented by some new features explained in Figure \ref{supp_figure_matrix}. Every stream corresponds to a line in the flowsheet matrix and is structured similarly: the vector $\bm{v}_i$ contains molar fractions, molar flowrate, mass flowrate, and the vector $\bm{y}$. $\bm{y}$ contains critical temperature, critical pressure, and the acentric factor for every component present in the flowsheet. Additionally, it contains the activity coefficients at infinite dilution for each binary subsystem present in the flowsheet. The vector $\bm{u}_i$ consists of a one-hot-encoding (OHE) for the connected unit, a variable for the continuous specification of this unit, and a binary number indicating if this unit requires a continuous specification or not. The vector $\bm{d}_i$ is a OHE, which describes the connectivity of the stream(s) that leave the unit connected in line $i$ (in a similar way as in \cite{goettl2021b}). The vector $\bm{m}_i$ consists of three binary indicators marking if stream $i$ is a feed stream, if the flowsheet synthesis is terminated, and if this line is used.

Additionally, one could add the temperature and pressure of the streams in the respective lines. However, as those are set to be constant within the example systems, this is omitted for now.

\section{Generation of problem instances}
\label{supp_section_sampling}

During training and evaluation, we set the molar flowrate of the feed stream for all instances to a constant value of $1\frac{\textnormal{\footnotesize Mmol}}{\textnormal{\footnotesize h}}$ (this value is chosen similarly to the processes examined in \cite{wang2018, kunnakorn2013, luyben2008, chen2015}). However, we note that the mass flowrate varies from instance to instance.

To generate a (random) problem instance, we sample the chemical system from $\{1, 2, 3, 4\}$ with a uniform distribution. To obtain the feed composition, we sample the molar fraction of the first component $x_1$ with a uniform distribution from $(0, 1)$. The molar fraction $x_2$ of the second component reduces to $x_2 = 1 - x_1$. 

\section{Parameters for \texorpdfstring{$\mathcal{C}_{\textnormal{\footnotesize literature}}$}{literature reward} and \texorpdfstring{$\mathcal{C}_{\textnormal{\footnotesize generic}}$}{generic reward}}
\label{supp_section_c_tables}

\begin{table}
	\caption{Parameters for $\mathcal{C}_{\textnormal{\footnotesize literature}}$. The base values for the investment costs of the unit operations are taken from \cite{chen2015} (for a mass flowrate $\dot{m}_{0}=25000\frac{\textnormal{\footnotesize kg}}{\textnormal{\footnotesize h}}$). Investment costs for all unit operations that are not listed are neglected. The prices for steam and components are chosen similarly as in \cite{goettl2021b}. For the sake of simplicity, we assume a constant price for all components except for the used solvents.}
	\begin{center}
		\begin{tabular}{|c|c|}
			\hline
			Parameter & Value \\
			\hline
			\hline
			$C_{0, \textnormal{\footnotesize distillation column}}$ / k\$ & 1000 \\
			\hline
			$C_{0, \textnormal{\footnotesize decanter}}$ / k\$ & 200 \\
			\hline
			$p_\textnormal{\footnotesize steam}$ / \$/kg & 0.04 \\
			\hline
			$p_\textnormal{\footnotesize feed component}$ / \$/kg & 0.5 \\
			\hline
			$p_\textnormal{\footnotesize solvent}$ / \$/kg & 0.05 \\
			\hline
		\end{tabular}
	\end{center}
\end{table}

\begin{table}
	\caption{Parameters for $\mathcal{C}_{\textnormal{\footnotesize generic}}$. As $\mathcal{C}_{\textnormal{\footnotesize generic}}$ assigns a score, the parameters do not have a monetary unit.}
	\begin{center}
		\begin{tabular}{|c|c|}
			\hline
			Parameter & Value \\
			\hline
			\hline
			$C_{\textnormal{\footnotesize distillation column}}$ & 10 \\
			\hline
			$C_{\textnormal{\footnotesize decanter}}$ & 5 \\
			\hline
			$C_{\textnormal{\footnotesize mixer}}$ & 0.5 \\
			\hline
			$C_{\textnormal{\footnotesize recycle}}$ & 0.5 \\
			\hline
			$C_{\textnormal{\footnotesize add solvent}}$ & 0.5 \\
			\hline
			$p_{\textnormal{\footnotesize feed component}}$ & 1000 \\
			\hline
			$p_{\textnormal{\footnotesize solvent}}$ & 100 \\
			\hline
		\end{tabular}
	\end{center}
\end{table}

\section{List of symbols}

\subsection*{Latin symbols}

\begin{tabbing}
	\hspace*{3cm}\= \kill 
	$a$ \> Hierarchical structured action\\
	$a_t$ \> Hierarchical structured action at timestep $t$\\
	$\mathcal A(s)$ \> Set of feasible actions at state $s$\\
	$\mathcal{C}_{\textnormal{\footnotesize generic}}$ \> Generic cost function\\
	$\mathcal{C}_{\textnormal{\footnotesize literature}}$ \> Literature cost function\\
	$C_{\textnormal{\footnotesize op}, u}$ \> Operating costs of unit $u$\\
	$C_u$ \> Fixed cost for unit $u$\\
	$c_o$ \> Gain from stream $o$ that leaves a process\\
	$d$ \> Dimension of latent space\\
	$\bm{e}$ \> One-hot-encoded vector\\
	$G(a)$ \> Gumbel noise for action $a$\\	
	$g_{\theta}$ \> Neural network\\
	$g^E$ \> Molar excess Gibbs energy\\
	$\bm{h}$  \> Linear combination of sequence output by MLP-Mixer block\\
	$\Delta h^{(m)}_{i, v}$ \> Enthalpy of evaporation for component $i$\\
	$H$ \> Affine embeding\\
	$k$ \> Beam width\\
	$L_i$ \> Granularity of the factorization at level 3b and level 3c\\
	$m$ \> Maximum for number of lines in flowsheet matrix\\
	$\dot{m}$ \> Mass flowrate\\
	$M_{s_t}$ \> Flowsheet matrix at state $s_t$\\
	$n$ \> Length of one line in the flowsheet matrix\\
	$\dot{n}$ \> Molar flowrate\\
	$N(s, a)$ \> Visit count of action $a$ at state $s$ in tree search\\
	$O$ \> Set of all leaving streams of a process\\
	$O_{\textnormal{\footnotesize spec}}$ \> Set of leaving streams that meet the purity specification of a process\\
	$p$ \> Price in in $\mathcal{C}_{\textnormal{\footnotesize generic}}$ and $\mathcal{C}_{\textnormal{\footnotesize literature}}$\\
	$\dot{Q}_\textnormal{\footnotesize Reboiler}$ \> Reboiler duty\\
	$Q(s, a)$ \> Estimated value for action $a$ at state $s$\\
	$\hat Q(s, a)$ \> Completed Q-value for action $a$ at state $s$\\
	$r$ \> Reward\\
	$\mathcal{R}$ \> Performance ratio\\
	$s$ \> State\\
	$s_t$ \> State at timestep $t$\\
	$t$ \> Timestep\\
	$U$ \> Set of unit operations used in a process\\
	$v$ \> Value, output of value head\\
	$\hat v(s)$ \> Interpolation of value of state $s$\\
	$\bm{w}_i$ \> Line $i$ of flowsheet matrix\\
	$x_i$ \> Molar fraction of component $i$\\
	$x^{(m)}_i$ \> Mass fraction of component $i$\\
	$z$ \> Number of actions sampled at root node during Sequential Halving\\
\end{tabbing}

\subsection*{Greek symbols}

\begin{tabbing}
	\hspace*{3cm}\= \kill
	$\alpha$ \> Lower bound of an interval describing a continuous action\\
	$\beta$ \> Upper bound of an interval describing a continuous action\\
	$\pi$ \> Policy, output by policy head\\
	$\hat \pi$ \> Improved policy\\
	$\theta$ \> Parameterization of neural network\\
	$\sigma$ \> Monotonically increasing map\\
\end{tabbing}

\section{List of abbreviations}

\begin{tabbing}
	\hspace*{3cm}\= \kill
	Ac \> Acetone\\
	AFS \> Automated flowsheet synthesis\\
	BS \> Beam search\\
	Be \> Benzene\\
	Bu \> Butanol\\
	Ch \> Chloroform\\
	Et \> Ethanol\\
	GNN \> Graph neural network\\
	ML \> Machine learning\\
	MLP \> Multilayer perceptron\\
	MCTS \> Monte Carlo tree search\\
	OHE \> One-hot-encoding\\
	PSE \> Process systems engineering\\
	Py \> Pyridine\\
	RL \> Reinforcement learning\\
	Te \> Tetrahydrofuran\\
	To \> Toluene\\
	Wa \> Water\\
\end{tabbing} 

\end{document}